\documentclass[11pt,twocolumn]{article}

\usepackage[a4paper,margin=1in]{geometry}

\usepackage{libertine}
\usepackage[libertine]{newtxmath}
\usepackage{titling}

\usepackage{amsmath,amsfonts}
\usepackage{graphicx}
\usepackage{booktabs}
\usepackage{multirow}
\usepackage{subcaption}
\usepackage{array}
\usepackage{enumitem}
\usepackage{placeins}
\usepackage{float}

\usepackage{algorithm}
\usepackage{algpseudocode}

\usepackage[numbers,sort&compress]{natbib}

\usepackage{hyperref}
\hypersetup{
    colorlinks=true,
    linkcolor=blue,
    citecolor=blue,
    urlcolor=blue
}
\usepackage[capitalize,noabbrev]{cleveref}

\title{
\vspace{-1.2em}
\hrule
\vspace{0.6em}
\centering
\Large\bfseries
Beyond the Performance Illusion:
Structure-Aware Stratified Partitioning and
Curriculum Distributionally Robust Optimization
for Spatially Correlated Domains
\vspace{0.6em}
\hrule
\vspace{0.8em}
}

\author{
\textbf{Prathamesh Patil}\textsuperscript{1} \quad
\textbf{Arpit Jain, PhD}\textsuperscript{1} \quad
\textbf{Aswanth Krishnan}\textsuperscript{1} \\[0.4em]
\textsuperscript{1}QpiAI India Pvt. Ltd. \\[0.3em]
\href{mailto:patil.p@qpiai.tech}{patil.p@qpiai.tech} \quad
\href{mailto:arpit.j@qpiai.tech}{arpit.j@qpiai.tech} \quad
\href{mailto:ashwanth.krishnan@qpiai.tech}{ashwanth.krishnan@qpiai.tech}
}

\date{}

\begin{document}
\twocolumn[
\maketitle
\vspace{-1.0em}
]

\begin{abstract}
Performance evaluation in AI systems commonly assumes that random dataset splits produce independent and identically distributed (i.i.d.) subsets. We show that this assumption often breaks down in spatiotemporally correlated domains such as aerial surveillance, precision agriculture, and medical imaging, leading to two systematic failures: \textbf{data leakage}, where correlated samples span training and validation splits and inflate performance estimates, and \textbf{hidden stratification}, where errors on minority subpopulations are obscured by aggregate metrics.

To address these issues, we propose a unified evaluation and training framework for spatially correlated data. We introduce \textbf{Structure-Aware Stratified Partitioning (SASP)}, which constructs validation splits that reduce spatiotemporal leakage while preserving meaningful class balance, and \textbf{Curriculum Distributionally Robust Optimization (CDRO)}, a curriculum-based relaxation of distributionally robust training that stabilizes optimization under these stricter splits.

Across multiple benchmarks, this combination yields consistently improved generalization, more reliable confidence calibration, and exposes failure modes that remain hidden under conventional random-split evaluation.
\end{abstract}

\section{Introduction}
The reliability of machine learning systems deployed in high-stakes environments critically depends on the validity of their evaluation protocols. In standard AI research, it is commonly assumed that randomly partitioning a dataset into training and validation subsets yields an unbiased estimate of generalization performance. This assumption relies on the premise that data points are independent and identically distributed (i.i.d.). While this approximation may hold for object-centric benchmarks such as ImageNet, where images are sourced from diverse and largely independent internet collections, it often breaks down in real-world data acquisition settings.

In many practical domains, data is collected sequentially or spatially, inducing strong spatiotemporal correlations between samples. Examples include aerial surveillance, where videos are captured by drones traversing urban environments, precision agriculture, where crops are imaged repeatedly across nearby plots, and medical imaging, where samples may be clustered by patient, scanner, or acquisition protocol. In such settings, random splitting can lead to evaluation protocols that substantially overestimate a model’s ability to generalize to genuinely novel conditions.

\subsection{The Failure of I.I.D. Assumptions}

We illustrate this issue using the VisDrone-DET benchmark \cite{zhu2018visdrone}, a widely used dataset for aerial object detection. VisDrone consists of video sequences collected during drone flights over urban scenes, where consecutive frames often depict the same physical environment under nearly identical viewpoints and lighting conditions. Under standard random splitting protocols, temporally adjacent frames may be assigned to different splits, placing near-duplicate samples in both training and validation sets.

This practice introduces \textbf{spatiotemporal leakage}: the model can achieve high validation performance by implicitly memorizing background structure rather than learning robust object-centric representations. As a result, performance measured on the validation set may fail to reflect behavior on truly unseen locations or acquisition conditions. Similar leakage patterns arise in other spatially correlated datasets, including precision agriculture benchmarks such as Global Wheat head detection and medical imaging datasets such as BCCD, where correlated samples are routinely split across evaluation folds.

\subsection{Hidden Stratification and the Long Tail}

A second, orthogonal challenge arises from the long-tailed nature of real-world data. Many datasets are dominated by common conditions, such as well-lit daytime scenes or typical anatomical presentations, while rare but critical subpopulations appear infrequently. These may include pedestrians under adverse lighting, crops affected by uncommon growth patterns, or pathological cases in medical imaging. As described by \citet{oakden2020hidden}, this phenomenon, termed \textbf{hidden stratification}, allows models to achieve strong aggregate performance while failing systematically on minority subgroups.

Random dataset partitioning exacerbates this issue by failing to control the distribution of these rare subpopulations across splits. In extreme cases, minority scenarios may be entirely absent from training data or insufficiently represented in validation sets, masking failure modes that only surface at deployment time.

\subsection{Overview of the Proposed Framework}

To address these challenges, we propose a unified framework that treats dataset partitioning and model training as coupled problems rather than independent design choices. First, we introduce \textbf{Structure-Aware Stratified Partitioning (SASP)}, a principled dataset splitting strategy that constructs validation folds which are semantically disjoint yet class-balanced, substantially reducing spatiotemporal leakage while preserving meaningful coverage of minority subpopulations. By design, SASP yields evaluation protocols that more faithfully reflect real-world generalization.

Second, we propose \textbf{Curriculum Distributionally Robust Optimization (CDRO)}, a training strategy designed to stabilize learning under these more stringent evaluation conditions. CDRO progressively emphasizes difficult or underperforming subgroups during training, mitigating the instability often observed when applying standard empirical risk minimization or fully adversarial robust objectives on rigorously partitioned data.

\subsection{Contributions}

In summary, our contributions are threefold:
\begin{enumerate}
    \item We identify and empirically characterize spatiotemporal leakage and hidden stratification as pervasive failure modes of standard random-split evaluation in spatially correlated computer vision datasets.
    \item We introduce Structure-Aware Stratified Partitioning (SASP), a dataset partitioning framework that enforces semantic disjointness while maintaining class balance, enabling more reliable evaluation of generalization.
    \item We propose Curriculum Distributionally Robust Optimization (CDRO), which improves training stability and model reliability under the rigorous evaluation protocols induced by SASP, and demonstrate its effectiveness across aerial surveillance, precision agriculture, and medical imaging domains.
\end{enumerate}

\section{Related Work}
\paragraph{Spatiotemporal Correlation and Evaluation Bias.}
The problem of evaluation bias induced by spatial and temporal autocorrelation has been extensively documented across ecology, climate modeling, and remote sensing. Early work by \citet{roberts2017cross} showed that random cross-validation can lead to severe underestimation of generalization error when samples exhibit spatial dependence, motivating spatial blocking, environmental clustering, and structured resampling strategies. Subsequent studies refined these insights, demonstrating that even weak autocorrelation can invalidate standard i.i.d.\ evaluation assumptions \citep{ploton2020spatial,wadoux2021spatialcv}.

Similar issues arise in modern computer vision benchmarks constructed from video streams, autonomous driving logs, medical imaging sequences, or geographically clustered imagery. Frame-level or instance-level random splits often leak near-duplicate content across training and evaluation sets, inflating reported performance and masking generalization failures \citep{torralba2011unbiased,azulay2019deep}. To mitigate this, prior work has proposed temporal splits \citep{karpathy2014large}, scene-level partitioning \citep{cordts2016cityscapes}, or location-based grouping \citep{sun2021scalability}. However, these approaches rely on hard heuristics and typically trade off semantic coverage, class balance, or evaluation realism.

Recent ICML and NeurIPS works have emphasized that evaluation protocol design is itself a critical source of inductive bias. For instance, \citet{gulrajani2021in} and \citet{taori2020measuring} showed that benchmark performance often fails to predict real-world robustness due to implicit correlations in dataset construction. Unlike prior approaches that impose coarse partitioning rules, our work formulates dataset splitting as a constrained optimization problem, jointly enforcing semantic disjointness and stratification. This enables evaluation folds that explicitly expose generalization gaps induced by spatiotemporal structure while preserving realistic semantic diversity.

\paragraph{Hidden Stratification and Long-Tailed Structure.}
Real-world vision datasets are inherently long-tailed, with rare but semantically important conditions underrepresented in aggregate metrics. \citet{oakden2020hidden} introduced the notion of \emph{hidden stratification}, demonstrating that models can achieve strong average performance while failing catastrophically on clinically or operationally critical subgroups. Follow-up work explored subgroup discovery \citep{chen2020failure}, slice-based evaluation \citep{koh2021wilds}, and reweighting or resampling strategies \citep{izmailov2022robustness} to expose and mitigate such failures.

However, these methods typically assume that evaluation splits already contain adequate representation of minority strata. When datasets exhibit spatiotemporal correlation, random partitioning can unevenly allocate rare scenarios, or eliminate them entirely from validation or test sets; thereby compounding hidden stratification effects. Recent ICML work on long-tailed and subgroup robustness has noted that evaluation bias can dominate training improvements when dataset structure is ignored \citep{creager2021environment,liu2023just}. Our structure-aware stratified partitioning (SASP) directly addresses this gap by ensuring systematic representation of minority semantic patterns across folds, enabling reliable subgroup-level evaluation without requiring explicit subgroup labels.

\paragraph{Distributionally Robust Optimization and Curriculum-Based Training.}
Distributionally Robust Optimization (DRO) provides a principled framework for improving worst-case performance under distribution shift. Methods such as Invariant Risk Minimization (IRM) \citep{arjovsky2019invariant}, Group DRO \citep{sagawa2019distributionally}, and related environment-based objectives aim to learn representations that generalize across predefined groups. While theoretically appealing, these methods often suffer from optimization instability, sensitivity to group definitions, and limited scalability to large vision models and object detection pipelines \citep{rosenfeld2021risks}.

Recent work has increasingly focused on stabilizing robust objectives via curriculum learning, adaptive reweighting, or gradual constraint enforcement, including large-scale methods for worst-case optimization \citep{levy2020large} and adaptive group discovery during training \citep{zhou2022modeling}. These approaches suggest that fully adversarial objectives may be unnecessarily brittle when evaluation protocols already induce strong distributional stress.

Our method builds on this insight by coupling SASP-induced evaluation rigor with a curriculum-based DRO formulation. Rather than enforcing worst-case constraints from the outset or relying on manually specified groups, we progressively increase robustness pressure as training stabilizes. This yields consistent gains under the more stringent, structure-aware evaluation regime introduced by SASP, while avoiding the optimization pathologies commonly observed in standard Group DRO for dense vision tasks.

\paragraph{Self-Supervised Representations for Semantic Structure.}
Self-supervised representation learning has emerged as a powerful tool for capturing semantic structure without task-specific supervision. Vision Transformers trained via contrastive or distillation-based objectives have been shown to encode global scene attributes such as texture, geometry, and illumination, enabling strong transfer across domains \citep{caron2021emerging,chen2021empirical}. Recent foundation models such as DINOv2 \citep{oquab2023dinov2} further demonstrate that self-supervised features can serve as reliable semantic descriptors across diverse visual distributions.

Prior work has primarily leveraged such representations as pretrained backbones or initialization schemes. In contrast, recent work has explored their use for dataset analysis and semantic shift characterization \citep{singh2022semantic}. Our work adopts this perspective by using self-supervised embeddings as a semantic metric for identifying latent correlation structure and guiding dataset partitioning. Crucially, this approach remains agnostic to downstream task labels and model architectures, allowing SASP to be applied uniformly across tasks while preserving evaluation fidelity.

\paragraph{Positioning of Our Work.}
Taken together, prior work has addressed spatiotemporal correlation, hidden stratification, robust optimization, and self-supervised representations largely in isolation. Our contribution lies in unifying these threads: we show that evaluation bias induced by structured data can be systematically exposed via SASP, and that curriculum-based DRO provides a practical and stable training response to the resulting distributional stress. Unlike prior DRO or robustness methods, our approach explicitly treats dataset partitioning and training dynamics as coupled design choices, yielding improvements that persist under realistic, structure-aware evaluation.

\section{Methodology}

Our framework consists of two sequential stages. First, we construct rigorous, leakage-resistant evaluation splits using \textbf{Structure-Aware Stratified Partitioning (SASP)}. Second, we train models under these stringent splits using \textbf{Curriculum Distributionally Robust Optimization (CDRO)}, which improves training stability and robustness when standard empirical risk minimization fails.

\subsection{Structure-Aware Stratified Partitioning (SASP)}

\subsubsection{Problem Setting}

Let $\mathcal{D} = \{(x_i, y_i)\}_{i=1}^N$ denote a dataset of images and labels. In some domains, additional metadata $m_i$ (e.g., sequence or acquisition identifiers) may be available. Our goal is to partition $\mathcal{D}$ into $K$ disjoint folds $\mathcal{F} = \{F_1, \dots, F_K\}$ such that evaluation faithfully reflects generalization under spatiotemporal correlation.

SASP enforces two complementary principles:
\begin{enumerate}
    \item \textbf{Structural Disjointness:} Highly correlated samples must not be split across folds.
    \item \textbf{Stratification:} Each fold should approximately preserve the global class distribution.
\end{enumerate}

When metadata is available, it is treated as a hard constraint; otherwise, structure is inferred entirely from visual semantics.

\subsubsection{Atomic Units and Structural Constraints}

We define \textit{atomic units} as the smallest indivisible groups of samples that must remain within the same fold. When metadata is available (e.g., video sequence identifiers), all images sharing the same metadata value form a single atomic unit. In the absence of metadata, each image initially forms its own atomic unit.

Atomic units serve as the fundamental entities for partitioning, ensuring that no explicit or implicit structural correlations are violated during splitting.

\subsubsection{Latent Semantic Clustering}

Metadata alone is insufficient to capture latent semantic correlations, such as visually similar scenes arising from different acquisition contexts. To capture such structure, we compute semantic embeddings using a frozen self-supervised vision model and represent each atomic unit by the mean embedding of its constituent images:
\begin{equation}
    z_j = \frac{1}{|U_j|} \sum_{x \in U_j} f(x),
\end{equation}
where $f(\cdot)$ denotes the embedding function.

We construct a similarity graph over atomic units using cosine similarity and identify connected components under an adaptive similarity threshold. Each connected component defines a \textbf{semantic cluster}, representing a latent neighborhood of visually correlated samples that must remain indivisible during partitioning.

\subsubsection{Hybrid Cluster-to-Fold Assignment}

Given a set of semantic clusters with heterogeneous sizes and class compositions, the goal is to assign each cluster to one of $K$ folds while maintaining stratification. This induces a constrained assignment problem balancing structural disjointness, fold capacity, and class distribution.

To achieve scalability, we adopt a hybrid strategy. Large clusters, which dominate fold statistics, are assigned first using a constrained optimization procedure that minimizes deviation from target class proportions. Smaller clusters are then greedily assigned to fill remaining capacity and correct residual imbalance. This two-stage procedure ensures stable stratification while preserving strict semantic disjointness.

\subsection{Curriculum Distributionally Robust Optimization (CDRO)}

Training under SASP splits is substantially more challenging than under random splits, as models can no longer exploit spurious correlations. In this setting, standard empirical risk minimization often exhibits unstable optimization and poor worst-case generalization.

CDRO addresses this by progressively emphasizing underperforming domains during training, approximating distributionally robust optimization through a curriculum-based relaxation.

\subsubsection{Curriculum Construction}

We partition training into $T$ consecutive phases. Each phase samples data from structurally defined groups (e.g., folds or semantic clusters) according to a distribution that reflects their relative difficulty. Group difficulty can be estimated using validation performance or embedding-based proxies, depending on domain and label availability.

At the end of phase $t$, we compute a difficulty score $\Delta_g^{(t)}$ for each group $g$, and update its sampling probability via:
\begin{equation}
    P_g^{(t+1)} = \frac{P_g^{(t)} \exp(\eta \Delta_g^{(t)})}{\sum_j P_j^{(t)} \exp(\eta \Delta_j^{(t)})},
\end{equation}
where $\eta$ controls the strength of reweighting. This biases training toward groups with poor generalization while avoiding fully adversarial objectives.

\begin{algorithm}[H]
\caption{Structure-Aware Stratified Partitioning (SASP)}
\label{alg:sasp}
\begin{algorithmic}[1]
\Require Dataset $\mathcal{D} = \{(x_i, y_i)\}_{i=1}^N$, number of folds $K$, similarity threshold $\tau$
\Ensure Fold assignment for all samples

\Statex
\State \textbf{Construct atomic units}
\If{metadata $m_i$ is available}
    \State Group samples with identical metadata into atomic units $\{U_j\}$
\Else
    \State Treat each sample as an atomic unit
\EndIf

\Statex
\State \textbf{Compute semantic representations}
\ForAll{atomic units $U_j$}
    \State Compute prototype embedding
    \[
    z_j \leftarrow \frac{1}{|U_j|} \sum_{x \in U_j} f(x)
    \]
\EndFor

\Statex
\State \textbf{Discover latent semantic clusters}
\State Build similarity graph $G$ over $\{z_j\}$ using cosine similarity
\State Connect $(i, j)$ if $S(z_i, z_j) > \tau$
\State Extract connected components $\mathcal{C} = \{C_1, \dots, C_M\}$

\Statex
\State \textbf{Assign clusters to folds}
\State Split clusters into large ($\mathcal{C}_{\text{big}}$) and small ($\mathcal{C}_{\text{small}}$)
\State Assign $\mathcal{C}_{\text{big}}$ via constrained optimization
\ForAll{$c \in \mathcal{C}_{\text{small}}$ (sorted by size)}
    \State Assign $c$ to the fold with the largest remaining capacity
\EndFor

\State \Return fold assignments
\end{algorithmic}
\end{algorithm}

\subsubsection{Final Stabilization Phase}

Strong reweighting can distort batch statistics and harm precision on common cases. To mitigate this, CDRO concludes with a stabilization phase in which sampling is returned to uniform, aggressive augmentations are reduced, and the learning rate is decayed. This allows the model to recover calibration and precision while retaining robustness learned during earlier phases.

---
\FloatBarrier
\section{Experiments}

We conduct an extensive empirical study across three spatially correlated object detection benchmarks to analyze the failure modes of standard evaluation protocols and to validate the effectiveness of our proposed framework. Our experiments are designed to answer the following questions:

\begin{enumerate}
  \item[(Q1)] To what extent do standard random splits induce spatiotemporal leakage and optimistic validation?
  \item[(Q2)] Can SASP recover meaningful latent domain structure without access to metadata?
  \item[(Q3)] Does SASP produce honest validation signals under strict structural constraints?
  \item[(Q4)] Can CDRO improve generalization once evaluation is made rigorous?
\end{enumerate}

\subsection{Datasets}

\textbf{Global Wheat Head Detection (GWHD)~\cite{david2020global}.}
GWHD contains wheat-field images collected from geographically distinct regions worldwide. Domain shift arises from differences in geography, climate, crop variety, and imaging protocols. We hold out Asian regions (UTokyo, NAU) as a fixed test set and treat all remaining regions as the training–validation pool. Importantly, no geographic metadata is provided to SASP.

\textbf{VisDrone-DET 2019~\cite{zhu2018visdrone}.}
VisDrone consists of 10,209 images extracted from 288 drone video sequences across 14 cities. Strong spatiotemporal correlation arises from adjacent frames, repeated camera trajectories, and shared urban environments.

\textbf{BCCD.}
BCCD is a medical microscopy dataset of blood cell images. Although smaller in scale, it exhibits strong acquisition-level correlations and serves as a representative safety-critical domain where validation reliability is crucial.

\subsection{Experimental Setup}

All experiments are implemented in PyTorch using the Ultralytics YOLO framework.

\begin{itemize}
    \item \textbf{Models:} YOLOv8n (lightweight), YOLOv5x and YOLOv10x (high-capacity).
    \item \textbf{Optimization:} SGD with momentum 0.937 and weight decay $5 \times 10^{-4}$.
    \item \textbf{Training Schedule:} Up to 200 epochs with early stopping (patience = 25).
    \item \textbf{CDRO:} Curriculum-based reweighting across SASP folds, followed by a final uniform, zero-augmentation phase.
    \item \textbf{Hardware:} NVIDIA L40S GPU (48 GB VRAM).
\end{itemize}

\subsection{Latent Domain Discovery without Metadata (Q2)}

We first evaluate whether SASP recovers meaningful latent domains. On GWHD, we cluster images using frozen DINOv2 embeddings and compare the resulting clusters with true geographic regions.

\begin{figure}[t]
    \centering
    \includegraphics[width=\linewidth]{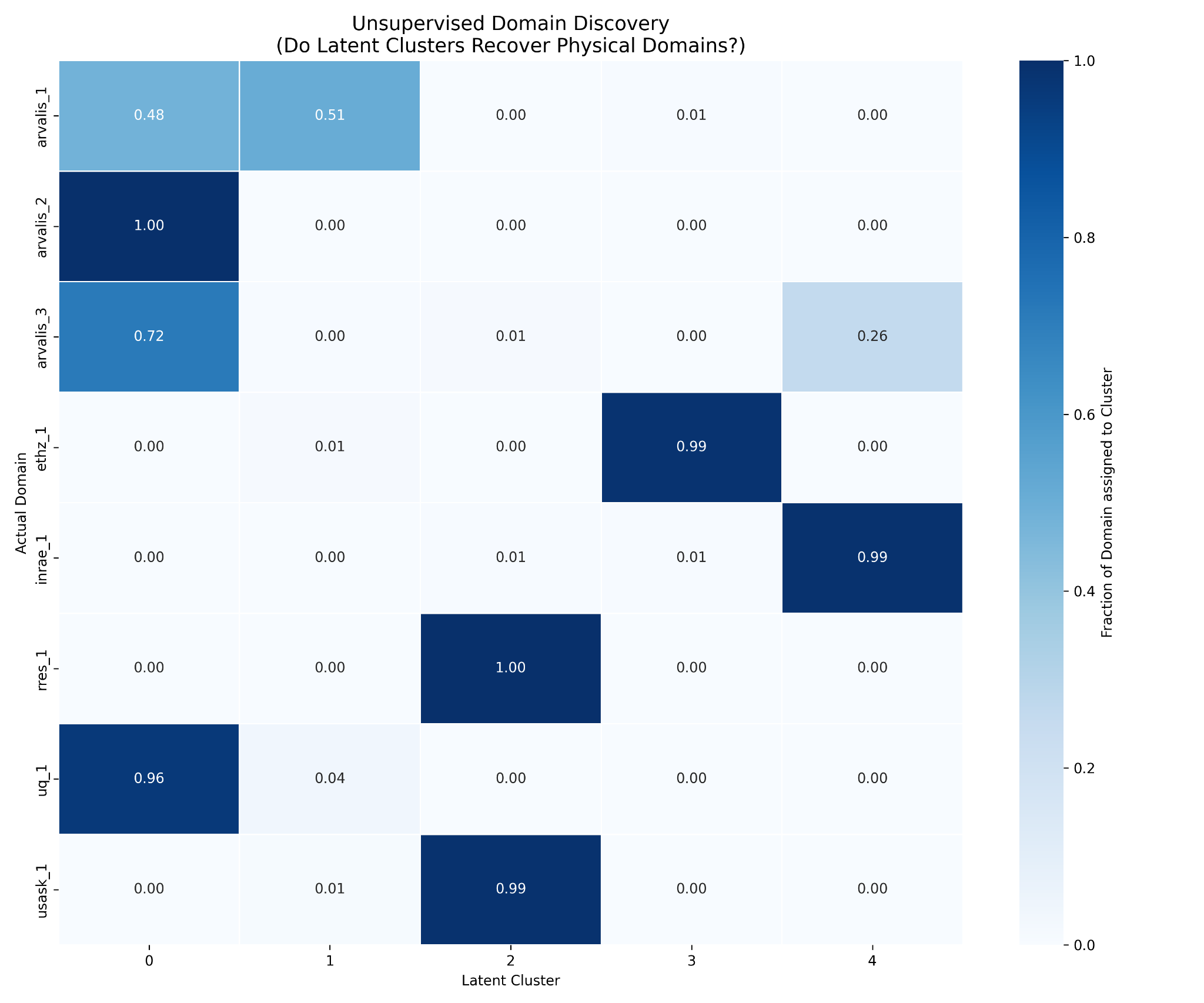}
    \caption{Unsupervised domain discovery on GWHD. Rows denote true geographic regions; columns denote latent clusters. Strong alignment indicates that SASP recovers physical domains without metadata.}
    \label{fig:wheat-domain}
\end{figure}

As shown in \cref{fig:wheat-domain}, latent clusters exhibit high purity with respect to geographic origin. This indicates that SASP discovers semantically coherent domains rather than relying on explicit metadata or heuristic grouping.

\begin{figure}[t]
    \centering
    \includegraphics[width=\linewidth]{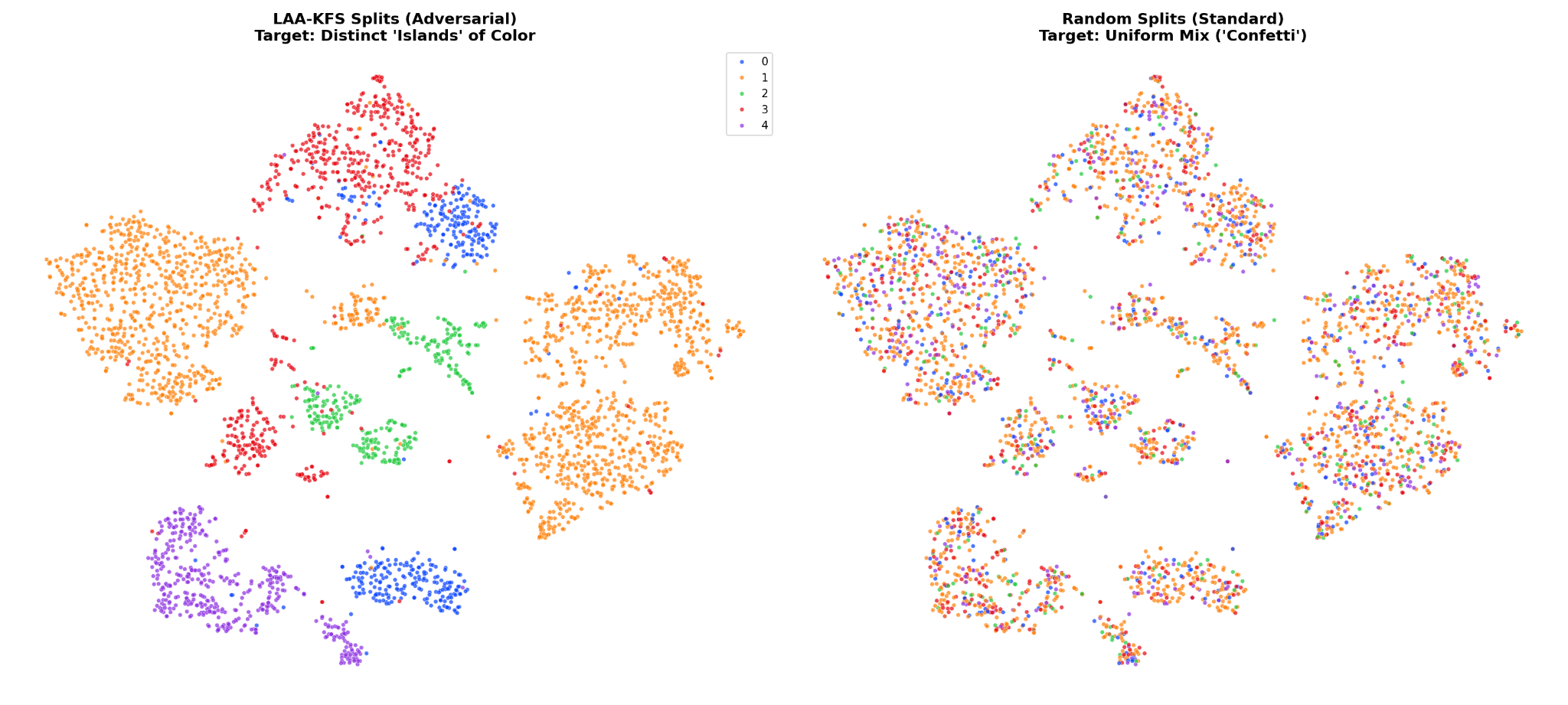}
    \caption{t-SNE visualization of GWHD embeddings. \textbf{Left:} SASP induces distinct semantic islands corresponding to latent domains. \textbf{Right:} Random splitting produces confetti-like mixing, obscuring domain structure.}
    \label{fig:tsne-wheat}
\end{figure}

The visualization in \cref{fig:tsne-wheat} provides qualitative evidence that SASP preserves global semantic structure, whereas random splitting actively destroys it.

\subsection{Quantifying Spatiotemporal Leakage (Q1)}

To measure leakage, we compute the maximum cosine similarity between each validation image and the training set using DINOv2 embeddings. Similarity above 0.95 indicates near-duplicate samples.

\begin{figure}[t]
    \centering
    \includegraphics[width=\linewidth]{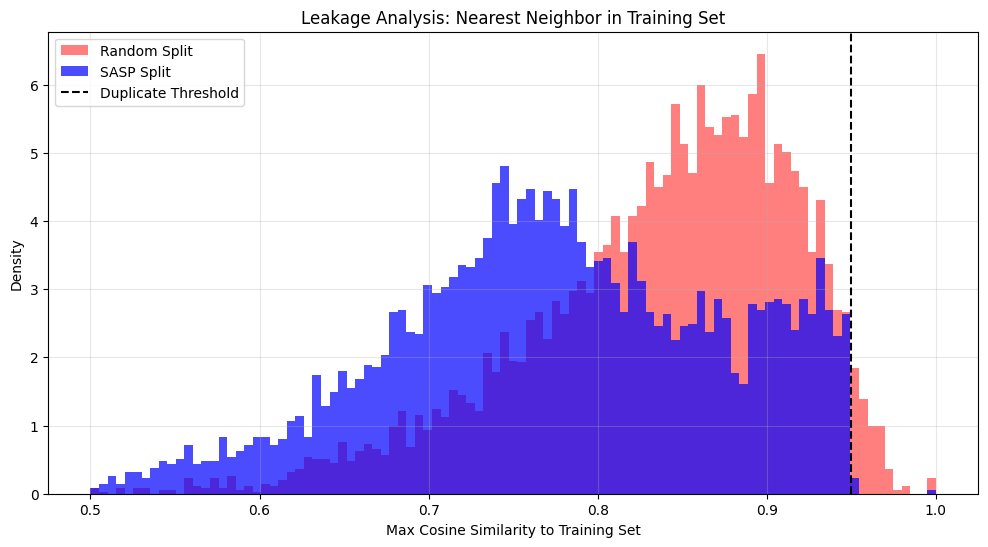}
    \caption{Nearest-neighbor similarity between validation and training images on VisDrone. Random splits exhibit substantial near-duplicate leakage, while SASP sharply reduces overlap.}
    \label{fig:leakage}
\end{figure}

As shown in \cref{fig:leakage}, random splits contain a significant mass of near-duplicate samples, explaining inflated validation performance. SASP reduces leakage by 98.5\%, creating a genuinely out-of-distribution evaluation.

\subsection{Structure vs. Stratification Trade-off}

A key challenge in strict partitioning is preserving class balance while enforcing structural constraints. To verify that SASP does not degenerate into naive blocking, we analyze class distributions across folds.

\begin{figure}[t]
    \centering
    \includegraphics[width=\linewidth]{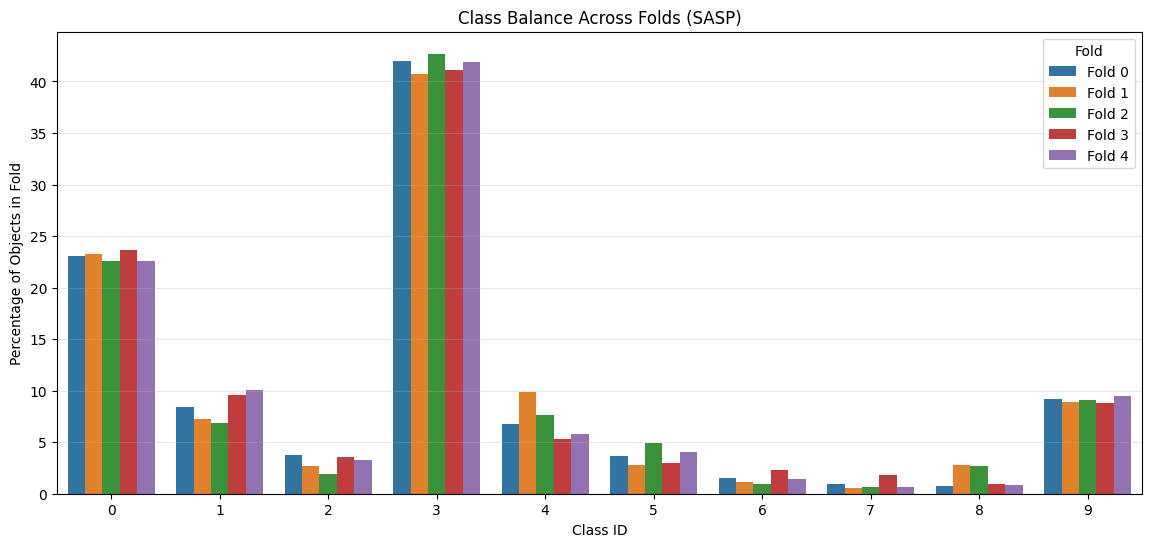}
    \caption{Class distribution across SASP folds on VisDrone. Despite strict structural constraints, SASP preserves class balance across folds.}
    \label{fig:class-balance}
\end{figure}

\cref{fig:class-balance} shows that SASP maintains low variance in class proportions, confirming that improvements are not due to favorable class reallocation.

\subsection{Evaluation Fidelity and Generalization under Structural Constraints (Q3, Q4)}

We now study detection performance under progressively stricter evaluation protocols:
(i) random splitting with empirical risk minimization (ERM),
(ii) SASP-based structural partitioning with ERM,
and (iii) SASP combined with curriculum distributionally robust optimization (CDRO).
This progression isolates the effect of evaluation fidelity from optimization strategy.
\begin{table}[t]
\caption{Random results are evaluated on leaky validation sets. GWHD reference performance is taken from the public winning solution~\cite{gwhd_winner}..
\textsuperscript{\dag} Proposed method.}
\label{tab:main-results}
\centering
\begin{small}
\begin{tabular}{llcc}
\toprule
Dataset & Method & Val mAP & Test mAP \\
\midrule
GWHD (v8n) & Random + ERM & 80.0 & 61.5 \\
           & SASP + ERM   & 70.0 & 62.5 \\
           & \textbf{SASP + CDRO}\textsuperscript{\dag} & \textbf{75.0} & \textbf{68.0} \\
\midrule
GWHD (v5x) & Random + ERM & 82.5 & 70.0 \\
           & \textbf{SASP + CDRO}\textsuperscript{\dag} & \textbf{77.0} & \textbf{72.0} \\
\midrule
VisDrone (v8n) & Random + ERM & 38.0 & 29.1 \\
               & SASP + ERM   & 30.6 & 29.2 \\
               & \textbf{SASP + CDRO}\textsuperscript{\dag} & \textbf{31.2} & \textbf{30.0} \\
\midrule
VisDrone (v10x) & Random + ERM & 56.0 & 50.0 \\
                & \textbf{SASP + CDRO}\textsuperscript{\dag} & \textbf{53.0} & \textbf{51.9} \\

\midrule
BCCD (v8n) & Random + ERM & 92.4 & 85.6 \\
           & SASP + ERM   & 87.2 & 87.5 \\
           & \textbf{SASP + CDRO}\textsuperscript{\dag} & \textbf{88.8} & \textbf{89.2} \\

\bottomrule
\end{tabular}
\end{small}
\end{table}

Three consistent trends are observed. First, random splitting yields overly optimistic validation estimates that fail to predict true test behavior. Second, enforcing structural constraints via SASP substantially reduces this mismatch, producing validation metrics that track test performance more faithfully. Third, once evaluation bias is removed, CDRO recovers generalization performance that would otherwise be obscured under leaky validation.

\paragraph{Integration of Latent Structure and Robust Optimization.}
Individually, the components underlying our framework—latent domain discovery, stratified partitioning, and distributionally robust optimization—have each been explored in prior work. However, these techniques are typically studied in isolation and under incompatible assumptions. In particular, most DRO formulations presuppose the existence of well-defined groups or environments, while existing domain discovery methods are rarely designed to produce partitions that are simultaneously semantically meaningful, stratified, and suitable for robust training.

Our contribution lies in unifying these elements into a coherent pipeline. SASP leverages self-supervised representations to uncover latent domain structure \emph{without metadata}, while explicitly enforcing stratification constraints that preserve class balance. These inferred domains then serve as well-formed groups for CDRO, enabling robust optimization in settings where group annotations are unavailable and naive clustering would lead to degenerate or unstable training. Empirically, we find that neither component is effective in isolation: latent structure alone yields pessimistic but underperforming models, while DRO without structurally grounded groups fails to stabilize. Their combination is therefore essential for achieving both evaluation fidelity and improved generalization.

\subsection{Training Dynamics and Validation Reliability}

To understand why CDRO is necessary, we analyze training dynamics on BCCD.

\begin{figure}[tb]
    \centering
    \includegraphics[width=\linewidth]{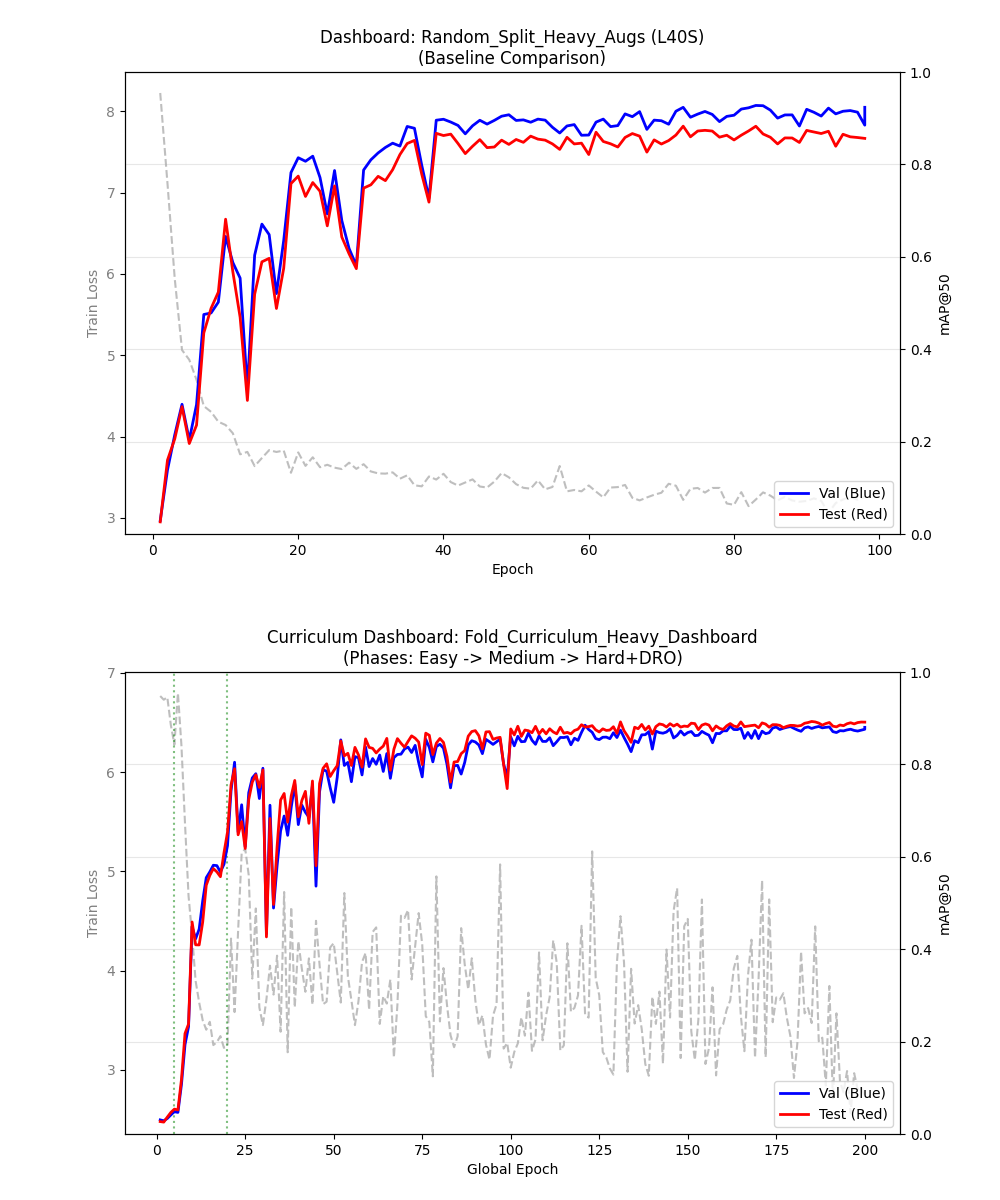}
    \caption{Training dynamics on BCCD with patience = 25. \textbf{Top:} Random splits exhibit a large validation–test gap, causing premature early stopping. \textbf{Bottom:} SASP + CDRO collapses the gap, restoring validation reliability.}
    \label{fig:training-dynamics}
\end{figure}

Under random splitting, validation peaks at 92.4\% while test performance remains at 85.6\%, causing early stopping to halt training prematurely. In contrast, SASP+CDRO aligns validation and test (88.8\% vs. 89.2\%), allowing early stopping to function as intended. This is shown in figure 5.

\subsection{Confidence and Calibration}

Finally, we analyze model confidence.

\begin{figure}[!t]
    \centering
    \includegraphics[width=0.9\linewidth]{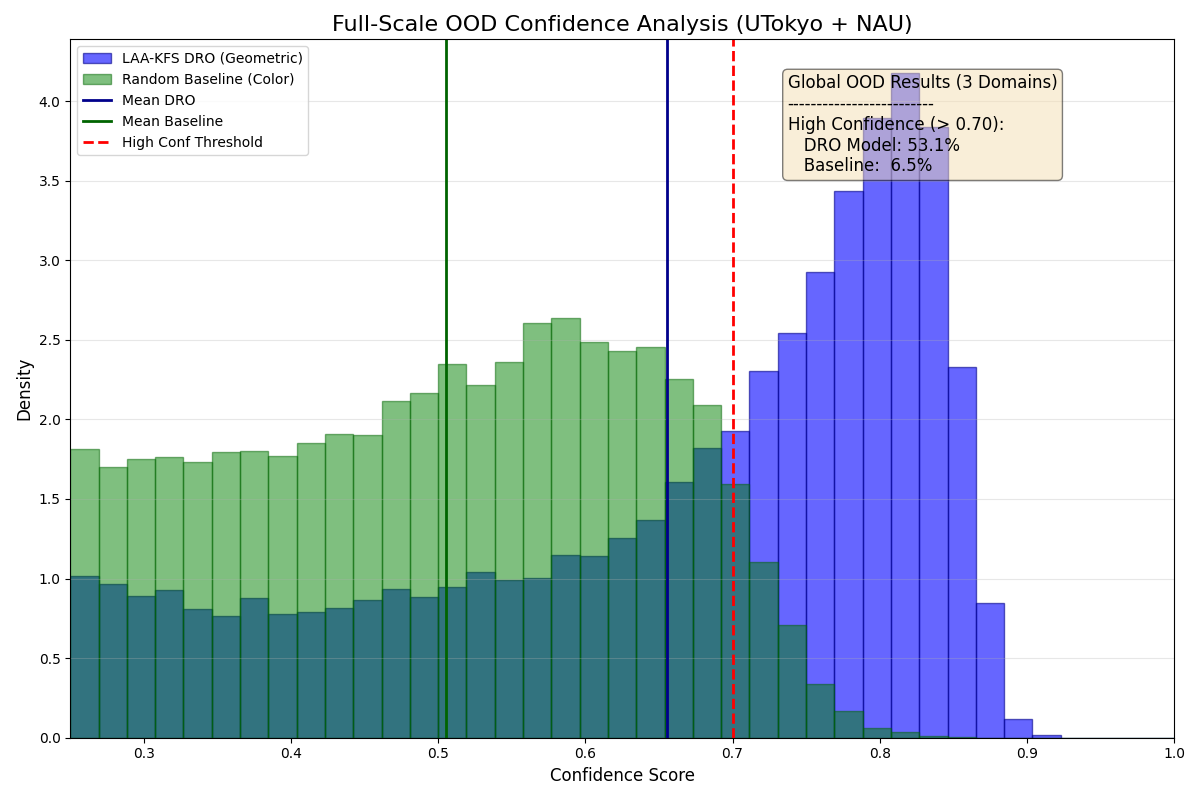}
    \caption{Prediction confidence distribution on GWHD. SASP + CDRO shifts predictions toward high-confidence regimes, indicating improved calibration.}
    \label{fig:confidence}
\end{figure}

As shown in figure 6, SASP+CDRO increases the fraction of predictions exceeding 0.7 confidence from 8\% to 53\%, indicating a fundamental shift in model behavior under distribution shift.

\FloatBarrier
\section{Conclusion}

This work exposes a fundamental weakness in how computer vision models are evaluated in spatially correlated domains. We show that standard random splitting protocols violate the i.i.d.\ assumption, inducing spatiotemporal leakage that inflates validation performance and obscures true generalization. 

To address this, we reframe dataset partitioning as a structured optimization problem. Structure-Aware Stratified Partitioning (SASP) enforces strict disjointness while preserving semantic and class balance, yielding validation sets that faithfully reflect out-of-distribution performance. We further demonstrate that once evaluation is made rigorous, standard ERM training becomes unstable, motivating Curriculum Distributionally Robust Optimization (CDRO), which restores validation reliability and improves generalization.

Across aerial surveillance, precision agriculture, and medical imaging benchmarks, our framework not only corrects misleading evaluation but also induces meaningful changes in training dynamics, confidence calibration, and early stopping behavior. These findings suggest that progress in robust perception requires rethinking evaluation protocols alongside model design.

Beyond computer vision, the issues identified in this work extend to other modalities with structured data collection. In large language models, for example, data contamination and benchmark leakage arise from near-duplicate documents, shared sources, and temporal overlap between training and evaluation corpora. Analogous to spatiotemporal leakage in vision, such correlations can inflate reported performance and obscure true generalization. We believe that structure-aware partitioning based on semantic similarity, combined with curriculum-based robust training, offers a promising direction for constructing more reliable evaluation protocols in language and multimodal models.

We believe future benchmarks should move beyond random splitting and adopt structure-aware evaluation to ensure that reported improvements correspond to real-world reliability.

\bibliographystyle{plainnat}
\bibliography{references}

\end{document}